\documentclass[10pt,journal,cspaper,compsoc]{IEEEtran}
\usepackage{graphicx}
\usepackage{comment}
\usepackage{amsmath,amssymb} 
\usepackage{color}

\usepackage{multirow}
\usepackage{paralist}

\usepackage[normalem]{ulem}
\usepackage{bbm}

\newcommand{\CUT}[1]{}
\newcommand{\ssum}{\mathrm{sum}}

\usepackage{wrapfig}

\begin{document}

\title{Fine-Grained Crowd Counting} 

\author{Jia~Wan, Nikil~Senthil~Kumar, and Antoni~B.~Chan 
	\IEEEcompsocitemizethanks{\IEEEcompsocthanksitem Jia Wan, Nikil Senthil~Kumar, and Antoni~B. Chan (corresponding author)  are with the Department of Computer Science, City University of Hong Kong.\protect\\
		E-mail: jiawan1998@gmail.com, nsenthilk2-c@my.cityu.edu.hk, abchan@cityu.edu.hk.
	}
	\thanks{}}

\markboth{Journal of \LaTeX\ Class Files,~Vol.~X, No.~X, XXX~XXXX}%
{Shell \MakeLowercase{\textit{et al.}}: Bare Demo of IEEEtran.cls for Computer Society Journals}


\IEEEcompsoctitleabstractindextext{
\begin{abstract}
	\CUT{Crowd counting from images has numerous practical applications, such as traffic control, ensuring public safety, and providing business insights. However, c}
	Current crowd counting algorithms are only concerned about the number of people in an image, which lacks low-level fine-grained information of the crowd.
	{For many practical applications, the total number of	people in an image is not as useful as the number of people
	in each sub-category. For example, knowing the number of people waiting inline or browsing can help retail stores;	knowing the number of people standing/sitting can help restaurants/cafeterias; knowing the number of violent/non-violent people can help police in crowd management.}	
	In this paper, we propose 
	fine-grained crowd counting, which differentiates a crowd into categories based on the low-level behavior attributes of the individuals (e.g. {standing/sitting} or violent behavior) and then counts the number of people in each category. To enable research in this area, we construct a new dataset 
	of four real-world fine-grained counting tasks: traveling direction on a sidewalk, standing or sitting, waiting {in line} or not, and exhibiting violent behavior or not. Since the appearance features of different crowd categories are similar, 
	 the challenge of fine-grained crowd counting is to effectively utilize contextual information to distinguish between  categories. 
	We propose a two branch architecture, consisting of a density map estimation branch and a semantic segmentation branch. 
We propose two refinement strategies for improving the predictions of the two branches.
First, to encode contextual information, 
we propose feature propagation guided by the density map prediction, which eliminates the effect of background features during propagation.
Second, 
we propose a complementary attention model to share information between the two branches.
Experiment results 
confirm the effectiveness of our method. 
\end{abstract}
}

\maketitle

\IEEEdisplaynotcompsoctitleabstractindextext
\IEEEpeerreviewmaketitle

\section{Introduction}
Crowd counting, which is defined as estimating the number of people in an image, has drawn increasing 
 attention because of its practical usage in public surveillance, traffic control and ensuring public safety, etc. For example, crowd count data in specific regions can be used to determine the departure intervals of public transportation for better traffic control, and to actively monitor the crowd sizes in public areas to avoid accidents that may occur due to overcrowding. The merits of crowd counting can also be extended to private surveillance 
 where business owners track the number of customers inside their stores to derive 
  business insights.

Although crowd counting is useful, it is solely concerned about the number of people in a crowd and, hence lacks information about the 
{underlying} behaviors of different groups within a crowd image, {e.g., the number of people waiting in line, the number of people sitting at tables, etc. 
Such {\em fine-grained} crowd information provides further useful details for crowd monitoring applications.}
{For example, by monitoring the number of people browsing, waiting in checkout lines, and entering/exiting a retail store, businesses can make staffing decisions to maximize productivity and minimize cost \cite{percolata}.}
Thus, in this paper,  we propose the problem of {\em fine-grained crowd counting} which differentiates the crowd into categories (e.g., waiting in line, standing vs.~sitting, violent vs.~non-violent), and estimates the spatial distribution and number of people in each category from the image (see 
Fig.~\ref{fig:cmp}).
{Note that many of these categories are defined by the {\em context} and {\em interaction} between individuals, and this context/interaction could take place over a long distance in the image.
For example, people waiting in a long line for a bus stand next to each other, and the line context is defined by the person standing at the head of the line next to the bus stop, {and goes from person to person until the last person in line (who may be on the other side of the image).}

\begin{figure}[t]
	\begin{center}
		\includegraphics[width=1.0\linewidth]{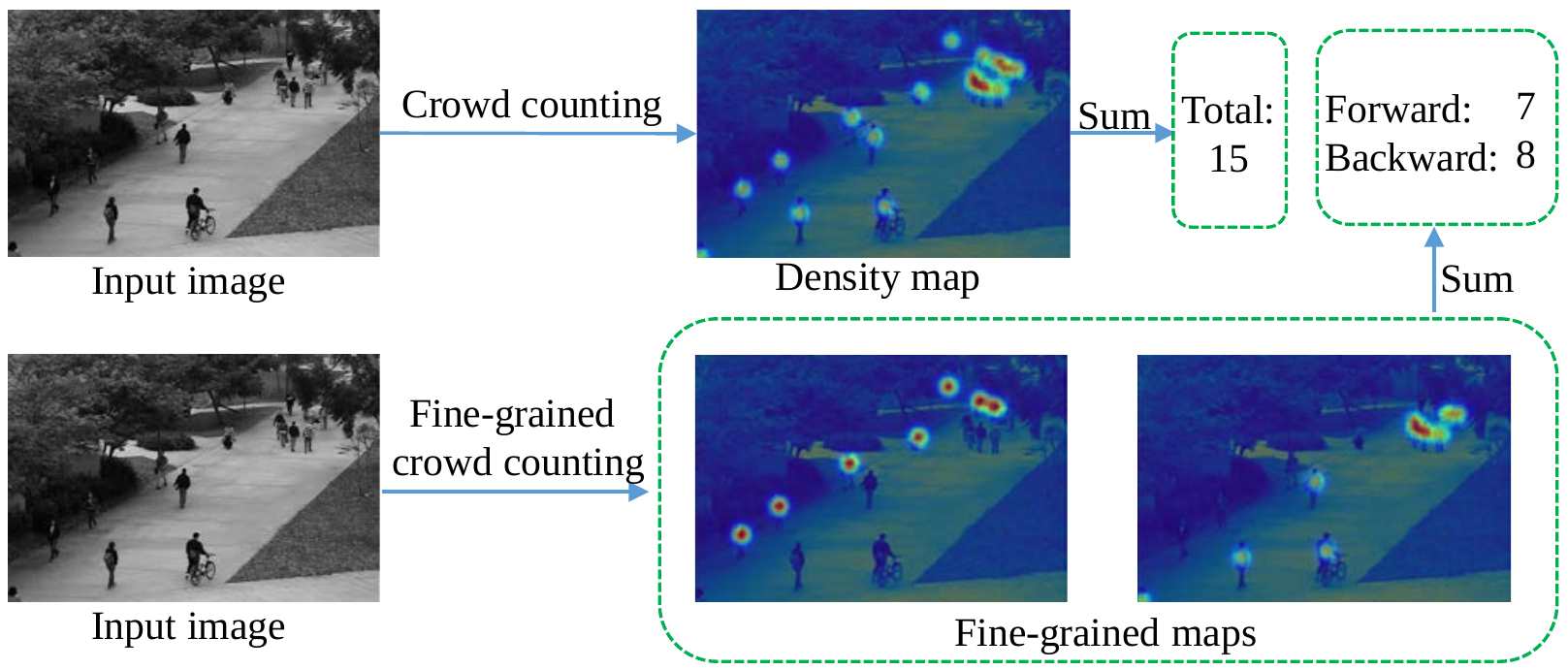}
	\end{center}
	\vspace{-0.5cm}
	\caption{\small {\em Crowd counting} counts the number of people in a whole image, while {\em fine-grained crowd counting} counts the number of people in different categories based on underlying behavior
	}\label{fig:cmp}	
\end{figure}

Compared to standard crowd counting, fine-grained crowd counting is a more challenging task because different categories will have similar appearance features, 
and thus context information needs to be used effectively (potentially over long ranges) to differentiate the categories.
{Hence, the straightforward approaches of using a separate crowd counting model for each crowd category or mixing the outputs of crowd-counting and semantic segmentation will not suffice to solve the problem effectively.}
To perform fine-grained counting, we propose a two branch {{\em coupled}} architecture, consisting of a density map estimation branch (for the whole crowd) and a segmentation branch (for categorization), whose outputs are mixed to obtain the category-specific crowd density maps.
{To share context information between the two branches,}
we propose two strategies to refine the predictions using complementary information from the other branch:
1) to utilize 
contextual information and capture high-order spatial relationships, we propose a density-aware feature propagation method for the segmentation branch, 
where the density map is used as guidance by decaying messages faster in low-density regions;
2) we propose a complementary attention module to utilize complementary information available between the two branches.
{Our models are general solutions, which are applicable to any  fine-grained counting task and can be easily extended to any number of crowd categories.}

For the experiments, we collect a new dataset for fine-grained crowd counting, containing  crowd images of people displaying different behaviors. The dataset contains four real-world tasks for identifying individual behaviors: walking towards vs.~walking away on a sidewalk; standing vs. sitting in a restaurant; waiting in line for the bus vs.~not waiting in line; violent behavior vs.~non-violent. {As a first work, we use those 4 real-world applications as examples since we believe those applications are practical and meaningful.}

In summary, the contributions of this paper are three-fold:

\begin{compactenum}
	\item We propose a novel problem, fine-grained crowd counting, which separates a crowd into categories and counts the number of people in each category. 
	\item We construct a new dataset consisting of four different applications of fine-grained crowd counting.
	\item To perform fine-grained counting, we propose a two-branch  coupled architecture with density-aware feature propagation, for incorporating long-range context information, and a complementary attention model.
\end{compactenum}

The remainder of this paper is organized as follows.  In Section \ref{text:related} we review the related works. In Section \ref{text:problem}, we define the fine-grained crowd counting problem and introduce the corresponding dataset.  In Section \ref{text:method} we present our two-branch coupled architecture, and in Section \ref{text:experiments} we present experiments. Finally, Section \ref{text:conc} concludes the paper.

\section{Related Works}
\label{text:related}

\begin{table*}[t]
	\begin{center}
	\caption{Fine-grained counting tasks in the proposed dataset}\label{tab:datasets}
		\begin{tabular}{ c  c  c  c  c  c  c c c} \hline
			Categories & Min. & Avg. & Max. & Total & \# of Image & \# of Class & \% of Class & Avg. Resolution \\ \hline
			Towards/Away & 13 & 27 & 52 & 65,259 & 2400 &2& 41, 59 & 158$\times$238 \\ 
			Standing/Sitting & 8 & 57 & 241 & 28,373 & 496 &2& 69, 31 & 787$\times$1199 \\ 
			Waiting/Not waiting & 2 & 22 & 65 & 9,311 & 420 &2& 50, 50 & 1080$\times$1920 \\ 
			Violent/Non-violent & 4 & 23 & 344 & 9,401 & 412 & 2& 37, 63 & 544$\times$838 \\ \hline 
		\end{tabular}
	\end{center}
\end{table*}
\begin{figure*}[t]
	\begin{center}
		\includegraphics[width=1.0\linewidth]{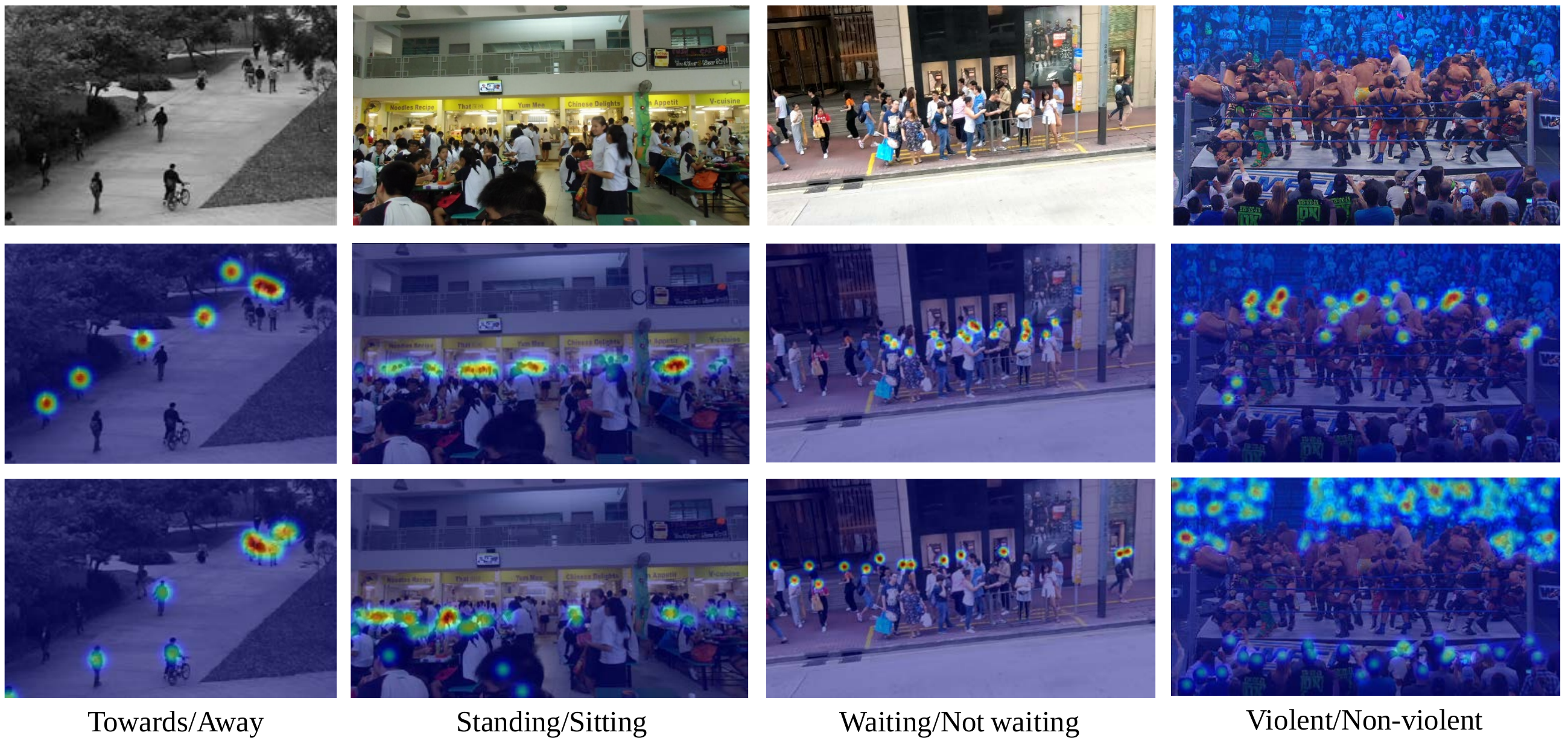}
	\end{center}			
	\vspace{-0.5cm}
	\caption{\small Example images and category density maps for the four tasks in the proposed fine-grained crowd counting dataset
	}\label{fig:datasets}
		\vspace{-0.5cm}
\end{figure*}

We review related works in crowd counting and  analysis.

\subsection{Crowd Counting}

Crowd counting algorithms aim to estimate the number of people in a whole image via either 
direct regression methods or density map based methods.
%
Traditional methods count the number of people by detecting individuals, either by their whole body or body parts \cite{4761705}. However, detection  approaches will fail on dense crowds with many occlusions.
To bypass 
detection, direct regression methods were proposed to directly predict the number of people from  low-level features 
\cite{4587569,5459191}. \cite{6619173} proposes to estimate the number of people using multiple features to improve counting performance. However, due to scale variation and occlusion, the performance of direct regression methods is also limited.

Density map approaches use a density map as an intermediate representation, which is then summed to get the count over a region. In recent years, density map methods achieve better performance than direct regression since the spatial distribution of the crowd is taken into consideration.
The typical design of density map estimation methods can be divided into two steps: 1) density maps are generated from dot map annotations;
2) deep learning models are designed to estimate density maps from images.
The most popular method to generate a density map is to convolve the dot map 
 with a fixed or adaptive Gaussian kernel. However, since hand-crafted density maps may not be optimal in the sense of end-to-end learning, \cite{wan2019adaptive} proposes a learning method to generate customized density maps for different networks/datasets. \cite{ma2019bayesian} proposes a Bayesian loss to evaluate the error between the dot ma and predicted density map.

{Various deep architectures for estimating density maps have been proposed to handle scale variations, refine results, perform domain adaptation, or exploit context information.}
To handle scale variation, \cite{7780439} proposes a multi-column network (MCNN) that extracts multi-scale features by three branches with different receptive field sizes and fuses the multi-scale features together, while \cite{switch} proposes to select between the branches. 
SANet \cite{cao2018scale} proposes a scale aggregation module applied to all convolutional layers, while \cite{babu2018divide} proposes a tree-structured CNN to deal with scale variations. \cite{DBLP:conf/bmvc/KangC18} uses an image pyramid to handle the different scale changes.
Density maps can be iteratively refined to obtain higher-quality predictions. 
\cite{ic-cnn}  proposes a two-stage approach where 
a low-resolution density map is refined into a high-resolution density map, 
 while \cite{topdown} uses  feedback 
and
\cite{spatial} proposes a region-based refinement method. 

Domain adaptation is essential for applying crowd counting technologies. 
 \cite{7298684} proposes to adapt a model to novel scenes by fine-tuning with a resampled dataset. \cite{wang2019learning} proposes a novel synthetic crowd dataset 
and a GAN-based algorithm to adapt from the synthetic dataset to real datasets.  \cite{ACNN} proposes an adaptive convolution that uses the camera parameters as side information.

To exploit {scene} contextual information, \cite{sindagi2017generating} proposes a contextual pyramid convolutional network (CP-CNN) that uses global and local context to improve counting performance, 
while \cite{xiong2017spatiotemporal} 
uses the temporal context in videos. 
\cite{liu2018leveraging} proposes a ranking-based method to use unlabeled data in learning.
A compositional loss is proposed in \cite{idrees2018composition} to solve global counting, density map estimation and localization together. Other works have shown that density maps can be applied to improve the detection and tracking performance in crowd scenes \cite{kang2018beyond,Ma_2015_CVPR,ren2018fusing}.
 A further survey of crowd counting algorithms can be found in \cite{sindagi2018survey}.

{Crowd counting algorithms are mainly concerned with counting the whole crowd, while in contrast fine-grained counting separates the crowd into groups and counts each group. While a crowd counting model could be learned for each group separately, this does not lead to good performance (see Sec.~\ref{text:experiments}) due to low appearance variations between groups and ignoring the long-range context. In this paper, we propose a novel architecture for fine-grained counting consisting of two-branches, density map estimation and category segmentation, as well as density-based feature propagation to better exploit long-range contextual information, and complementary attention to mix information between branches.
}

\subsection{Crowd Analysis}
Previous works on crowd analysis focus on classifying the crowd using high-level attributes, such as 
collectiveness, cohesiveness, and density level 
\cite{7434020}. However, low-level individual behavior properties, such as facing direction, standing, sitting, are not considered. \cite{shao2016learning} proposes more properties of crowds such as stability, uniformity, and conflict which are mainly determined by the crowd motion.
\cite{dupont2017crowd} proposes to classify crowd images by different crowd flow properties, while \cite{hassner2012violent} proposes 
classifies whether a crowd image contains violent behavior or not. 
Detection of social interactions is considered in \cite{zanlungo2013walking}, but is only concerned with small groups (less than 4 people). 
Although crowd analysis algorithms can be used to distinguish different crowd types, they are mainly concerned with the high-level attributes of the crowd. 
In contrast to these works, the proposed fine-grained crowd counting aims to find the spatial distribution and underlying behavior properties of the crowd.
%
%
%
Finally, general object detection \cite{7485869,YOLO} also focuses on detection of individuals, but has difficulty distinguishing individual's behavior since the inter-category appearance variation is low.
Furthermore detection methods  ignore the relationship and interaction between individuals in the crowd,  {thus are not able to capture the context information necessary for fine-grained counting.}


\section{Problem Definition and Dataset}
\label{text:problem}

\begin{figure*}[t]
	\begin{center}
		\includegraphics[width=\linewidth]{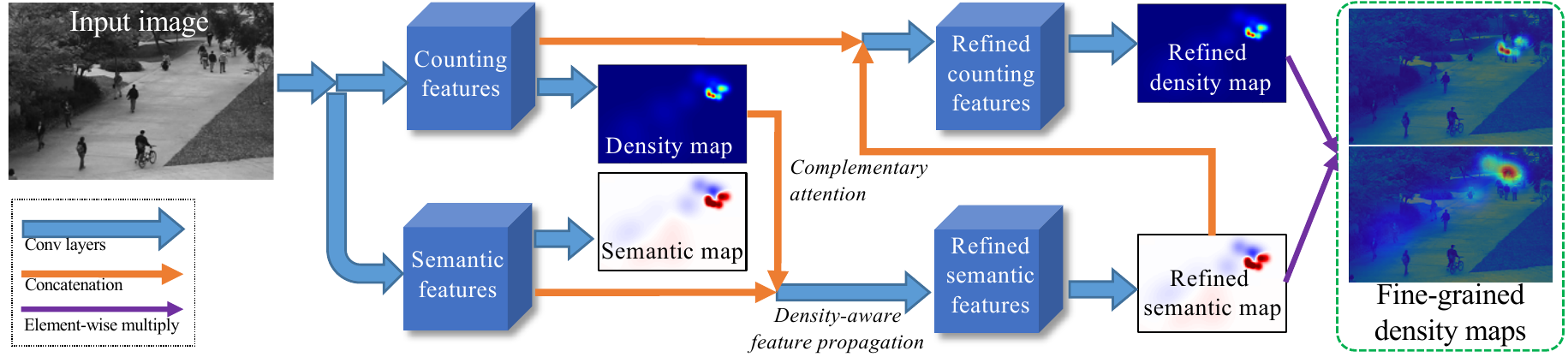}
	\end{center}

	\caption{
	{Our proposed architecture for fine-grained crowd counting uses two coupled branches to predict an overall crowd density map and a segmentation map. Information between branches are shared to refine the initial predictions. Density-aware feature propagation incorporates long-range context information, while complementary attention refines semantic features using the density map, and vice versa.
	The fine-grained density maps are obtained by multiplying the predicted density map with the predicted segmentation map for each category.
	}
	}\label{fig:pipeline}

\end{figure*}

%
The goal of fine-grained crowd counting is to divide the crowd into categories and then estimate the spatial distribution and  number of people in those categories. 
{Our problem is mainly formulated through a set of task-specific applications, which are generally orthogonal to each other.}
 For example, we want to count the number of people  waiting in line for a bus versus the number of people walking by, or the number of people sitting in a cafeteria versus the number of people standing.
{Thus, it is not necessary or plausible to need to solve all the tasks at the same time, since mixed classes are usually meaningless (e.g., a person who is sitting, and waiting for a bus, and fighting at the same time).}

Previous crowd counting methods can be used to obtain the overall crowd density map, 
and thus the main challenge of fine-grained crowd counting is effectively classifying crowd categories. Since people within the same category have certain relationships with each other, the key is to exploit contextual information to model that relationship (e.g., people waiting for a bus may stand in a line next to each other), 
and this context occurs over long-ranges in the image (e.g., the line is long, and the first person in line stands next a bus stop sign).
{In another example, for the violent/non-violent task, the posture and spacing of people, and their interaction, can give clues about the crowd class.}
Therefore, 
encoding the contextual information is crucial for fine-grained crowd counting.

\subsection{Fine-Grained Crowd Counting Dataset}

To enable research on fine-grained crowd counting, we 
construct a new dataset with four fine-grained counting tasks, based on real-world requirements:
1) walking towards or away on a sidewalk; 
2) standing or sitting;
3) waiting for bus or not waiting;
4) violent actions or non-violent.
The images for the first task are %
from the UCSD crowd  dataset \cite{4587569}, while the images for the fourth task are from 
 a violence detection dataset \cite{hassner2012violent} and a Kaggle dataset\footnote{https://www.kaggle.com/mohamedmustafa/real-life-violence-situations-dataset}. 
We collected the images for tasks 2 and 3 from the Internet.
{To increase dataset diversity and evaluate algorithms under different image qualities, the dataset is composed of images with a variety of resolutions, use different color spaces (grayscale vs. RGB), and are captured from different sources.}

Similar to crowd counting, the images were labeled with dot annotations indicating each person's location in the image. 
In addition, %
each dot was also assigned a crowd category for fine-grained counting.\footnote{{For the Towards/Away and Waiting/Not-waiting tasks, the raw video source was used for annotating the classes to reduce ambiguity.}}
{Since the tasks are orthogonal, we only label the classes specific for each task.}
 Information about the dataset and typical images with annotations are shown in Table \ref{tab:datasets} and Figure \ref{fig:datasets}.
 
 To better understand the proposed dataset, we show the spatial probability maps of each class in Figure \ref{position}.  We find that the spatial probability maps of each class are overlapping, making it impossible to distinguish different classes by a simple region-of-interest assignment map.  Also, immediate adjacency to a point-of-interest (e.g., bus-stop) is not enough to predict the class,  since  behaviors can be defined by  person-to-person context.
{For example, in Figure \ref{fig:datasets} (3rd column), the bus-stop is on the right, and the line context moves to the left away from the bus-stop).}
 
 We also visualize the average images for different tasks in Figure \ref{background}. For each task, the scenes are of similar context (e.g., street scenes), since they are task-specific.
 Towards/Away has a simple background due to the static camera. For other tasks, the background varies significantly.
 
 \begin{figure}[tb]
 	\scriptsize
 	\begin{center}
 		\begin{minipage}[t]{0.7cm}\centering\vspace{-3.5cm}$p_1$\vspace{1.1cm} \\ $p_2$ \vspace{1cm} \\ $\log \frac{p_1}{p_2}$\end{minipage}
 		\includegraphics[width=0.9\linewidth]{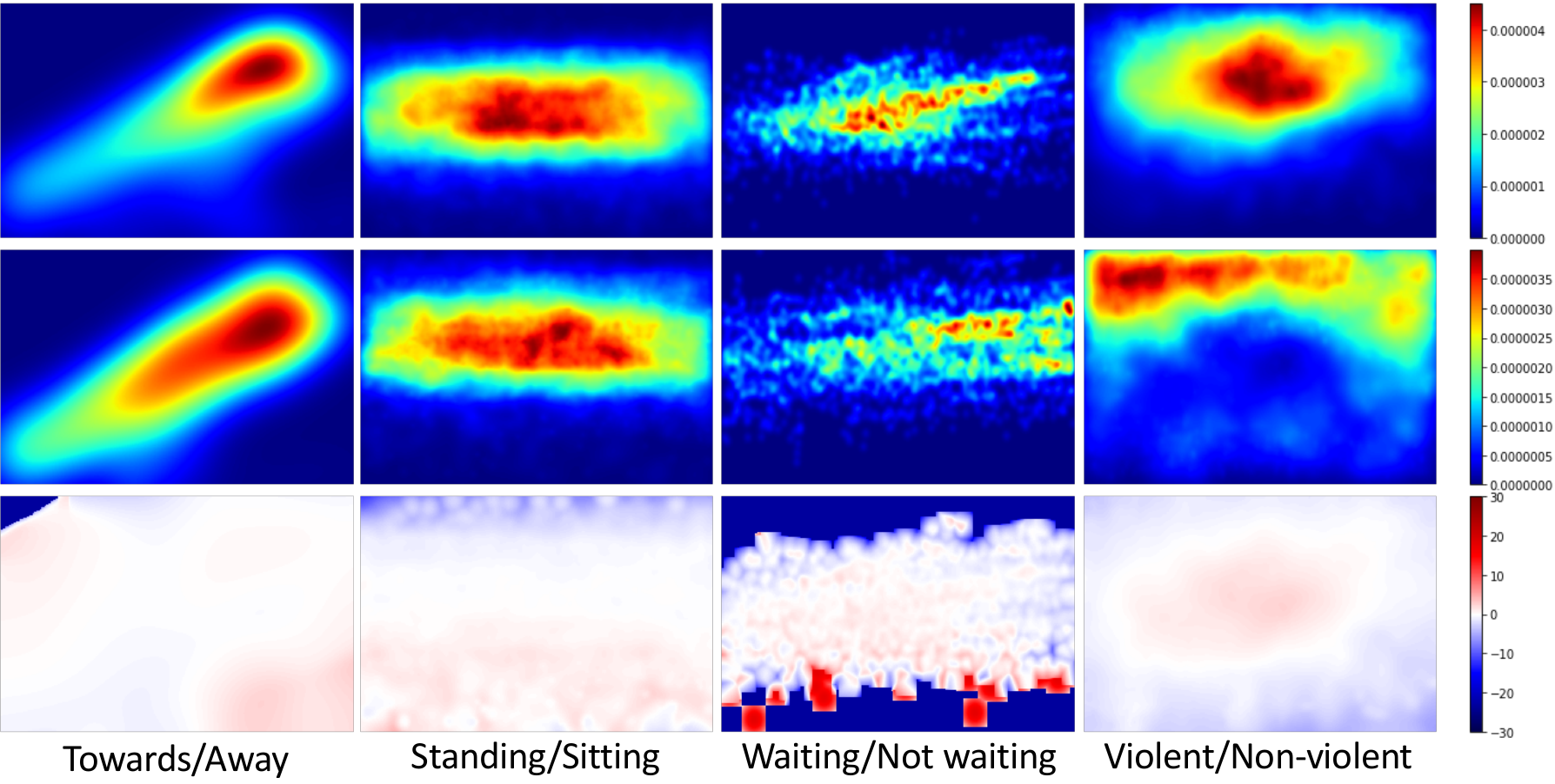}
 	\end{center}
 	\caption{(top and middle) probability map of class 1 or 2. (bottom) $\log (p_1/p_2)$: white indicates equally likely to be either class, while red or blue indicate higher probability of class 1 or class 2}\label{position}
 \end{figure}
 
 \begin{figure}[tb]
 	\begin{center}
 		\includegraphics[width=\linewidth]{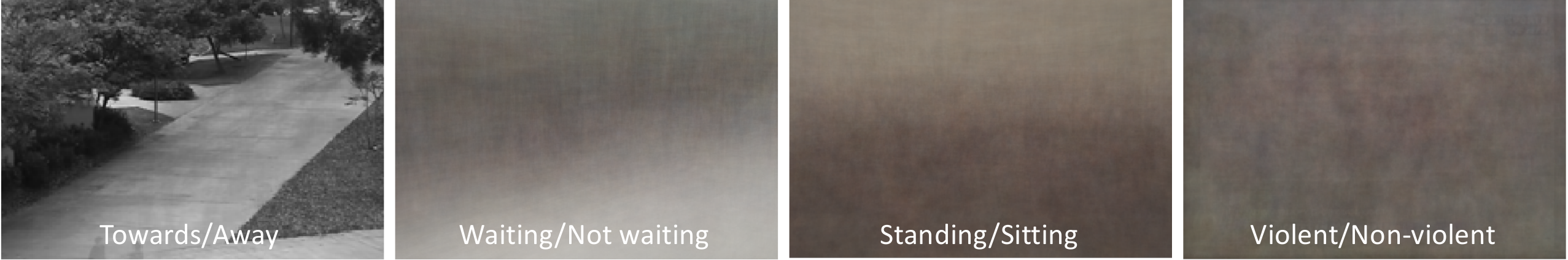}
 	\end{center}
 	\caption{Average images for different tasks. Tasks 2-4 are blurry, indicating non-static backgrounds
 	}\label{background}
 \end{figure}


\subsection{Ground-truth Density and Segmentation Maps}
We next discuss how to generate the ground-truth (GT) density and segmentation maps for training fine-grained crowd counting models. Let there be $K$ crowd categories indexed by $j\in\{1,\cdots,K\}$.
Density maps are generated from a dot map by convolving with a Gaussian kernel.
Specifically, given an image $I$ with width and height $(w,h)$ and the corresponding dot map annotation  for the $j$th crowd category $D_j\in \mathbb{R}^{h\times w}$, the corresponding density map $Y_j\in \mathbb{R}^{h\times w}$ for the $j$th category is
%
$Y_j=D_j*k_{\sigma}$,
where $k_{\sigma}$ is a 2D Gaussian kernel with bandwidth $\sigma$ and $*$ is  convolution.
The  kernel bandwidth can be fixed or change with pixel location, based on perspective \cite{7780439} or crowdedness \cite{7780439}. 
 The generation of the density map is equivalent to placing a Gaussian on each dot annotation.
\CUT{,
\begin{equation}
Y_j(p)=\sum_{p' |D_j(p')=1} \mathcal{N}(p | p', \sigma^2 I),
\end{equation}
where $p$ and $p'$ are pixel locations, and $\mathcal{N} (p | \mu,\Sigma)$ is a multivariate Gaussian with mean $\mu$ and covariance $\Sigma$. 
}

{The GT category segmentation maps are generated from the GT  density maps.}
First, the background segmentation map $S_{K+1}$ 
of low density regions is
\begin{equation}
S_{K+1} = \mathbbm{1}({\sum_{j=1}^K Y_j \leq \epsilon)},
\end{equation}
where $\epsilon$ is a low-density threshold, and $\mathbbm{1}$ is the indicator function.
%
The soft segmentation ground-truth for the $j$th category $S_j\in \mathbb{R}^{h\times w}$ is 
\begin{equation}
S_j = \frac{Y_j}{\eta+\sum_{j'=1}^{K} Y_{j'}},
\end{equation}
{where $\eta$ is a small number to prevent dividing by 0}.
{Note that the GT 
 segmentation maps are rough segmentations (not per-pixel GT as in semantic segmentation) since they are generated only from the density maps.  For fine-grained crowd counting, this rough segmentation is sufficient since the main goal is to assign each non-zero density pixel to a category, rather than all the image pixels.}


\section{Methodology}
\label{text:method}

{We propose a two-branch coupled architecture with density-aware feature propagation and complementary attention for fine-grained crowd counting.}
The overall pipeline of the proposed method is shown in Fig.~\ref{fig:pipeline}.
%
The architecture contains two branches; 
the density map branch predicts the density map of the overall crowd, while the segmentation branch classifies pixels into categories. 
The two branches share information to refine each other's predictions.

\begin{table}
\caption{FCN layer settings for density map estimation and semantic segmentation. 
ConvX-Y indicates convolution layer with filter size X with Y output channels}
\small
\centering
\begin{tabular}{|c|l|l|}
\hline
 & \multicolumn{2}{l|}{\underline{Shared Feature Extractor}} \\
Layer 1 & \multicolumn{2}{l|}{Conv5-16 + LReLU(0.1)} \\
Layer 2 & \multicolumn{2}{l|}{Conv5-16 + LReLU(0.1) + MaxPool} \\ 
Layer 3 &  \multicolumn{2}{l|}{Conv5-32 + LReLU(0.1)} \\
Layer 4 &  \multicolumn{2}{l|}{Conv5-32 + LReLU(0.1)+ MaxPool} \\
Layer 5 &  \multicolumn{2}{l|}{Conv5-64 + LReLU(0.1)} \\
\hline 
& \underline{Density Estimation} & \underline{Segmentation} \\
Layer 6 & Conv5-32 + LReLU(0.1) & Conv5-32 + LReLU(0.1)  \\
Layer 7 &  Conv5-1 & Conv5-K + Softmax \\
\hline
\end{tabular}
\label{tab:FCNs}
\end{table}

Specifically, the first-stage density map $\hat{Y}'$ and segmentation map $\hat{S}'$ are predicted using two { 7-layer Fully Convolutional Networks (FCNs) with 5 shared feature extraction layers (see Table~\ref{tab:FCNs}).} 
{We can also use other backbone feature extractors, such as CSRNet \cite{li2018csrnet}.}
Next, the segmentation feature map is iteratively propagated to encode contextual information, 
with the propagation guided by the first-stage density map $\hat{Y}'$.
Since features from background (non-crowd) pixels may mislead the classification of foreground pixels, those background features should decay faster than the foreground features. 
{To focus the refinement stage on important regions, we also use a complementary attention method where the density map is concatenated with the segmentation features to predict a refined segmentation map $\hat{S}_j$, and vice versa, the segmentation map is concatenated with the density features to predict a refined density map $\hat{Y}$.}
%
Finally, the predictions of the fine-grained density maps are formed by element-wise multiplying the overall  density map and the segmentation map for each category,
	\begin{align}
	\hat{Y}_j = \hat{S}_j \odot \hat{Y}
	\end{align}
where $\hat{Y}_j$ is the predicted density map for category $j$.

\subsection{Density-aware feature propagation}
To encode contextual information for better segmentation of fine-grained classes, we propose a density-aware feature propagation method for the segmentation branch.
{The aim is to iteratively propagate features in the feature map along high density regions to encode context and generate refined maps.}
{Note that semantic segmentation methods also use spatial smoothing of feature maps \cite{krahenbuhl2011efficient,toyoda2008random}. 
However, 
in contrast to standard semantic segmentation, which applies the spatial smoothing uniformly, our density-aware propagation applies smoothing only on regions connected through high crowd density values, so that these context features do not affect people who are not connected in a crowd.}

We consider three potential models for feature propagation: Conditional Random Fields (CRFs) \cite{zheng2015conditional}, Graph Convolutional Network (GCN) \cite{defferrard2016convolutional}, and Stacked Hourglass \cite{newell2016stacked}:

\begin{compactitem}
\item \uline{CRF} is a graphical model that is widely used in semantic segmentation to improve accuracy. The relationship between pixels is modeled with probabilistic inference under the 
assumption that similar pixels should have similar labels. 
Given an input image $I$ which has $n$ pixels, the segmentation of $I$ can be modeled as a random field $S= (S_1,S_2,\dots, S_n)$ with conditional distribution 
\begin{align}
p(S|I) = \tfrac{1}{Z(I)} \exp(-E(S|I)),
\end{align}
where the energy $E(S|I)$ is
\begin{equation}
E(S|I) = \sum_i \psi_u(S_i) + \sum_{i < j} \psi_p(S_i, S_j).
\end{equation}
$\psi_u(S_i)$ is the unary potential that measures the negative log-likelihood of assigning pixel $i$ to segment $S_i$, and $\psi_p(S_i,S_j)$ is the pairwise potential of assigning pixels $i$ and $j$ to segments $S_i$ and $S_j$. Note that, message passing is performed in local regions and hand-crafted features are used for evaluating the similarity between pixels, as in  \cite{zheng2015conditional}.


\item \uline{GCN} is a generalization of CNNs to graph structures, which are used to exploit contextual information in segmentation \cite{zhang2019dual}. Given an image $I$ with $n$ pixels, the adjacency matrix between pixels $A\in \mathbb{R}^{n\times n}$ is defined by the cosine similarity between their corresponding 
deep features $F\in \mathbb{R}^{n\times d}$, 
where $d$ is the feature dimension. The feature propagation is defined as:
\begin{equation}
H^{l+1}=\sigma(AH^lW^l),
\end{equation}
where $H^l$ is the output of the $l$-th layer,  $H^0=F$, and $\sigma(\cdot)$ is a non-linear activation, and $W^l$ is the weight matrix of the $l$-th layer. The 
propagation is 
performed in local regions. 

\item \uline{Stacked Hourglass} is an iterative bottom-up, top-down model that combines global and local contextual information. After the initial features are extracted, the features are iteratively refined using an hourglass module, as shown in 
Fig.~\ref{fig:hourglass}.  See \cite{newell2016stacked} for more details.
\end{compactitem}
\begin{figure}
	\begin{center}
		\includegraphics[width=0.8\linewidth]{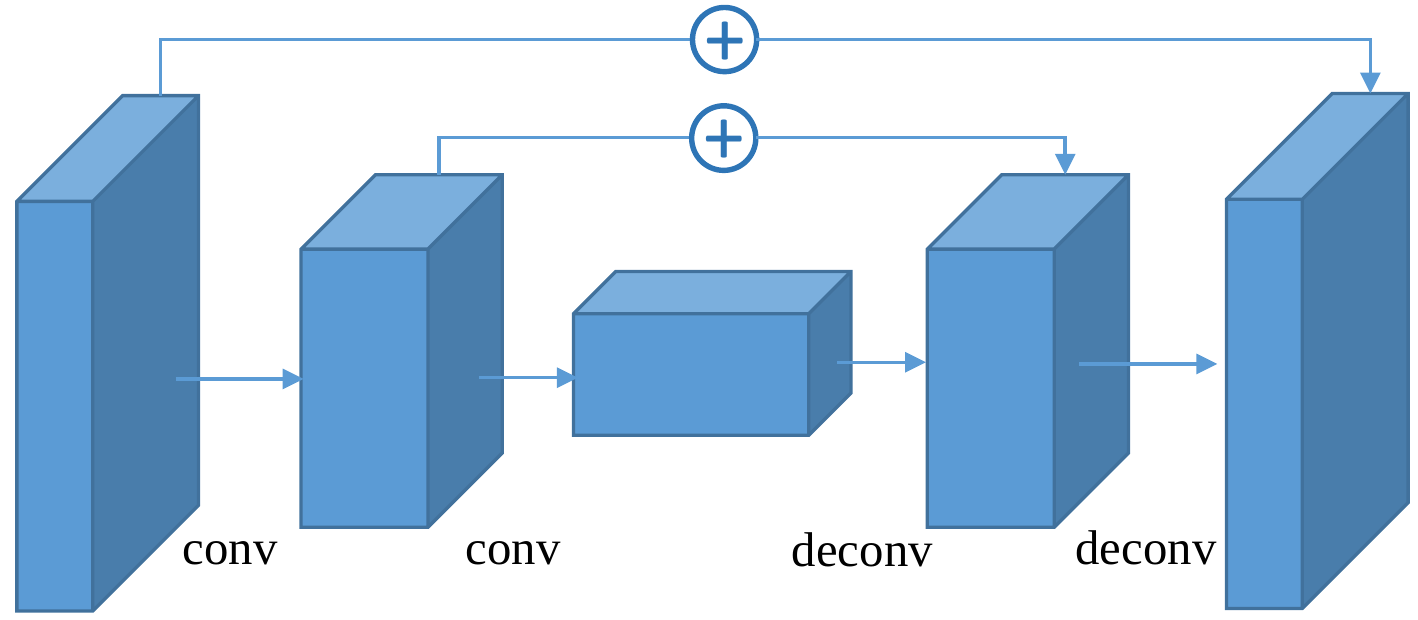}
	\end{center}
	\caption{\small Architecture of one hourglass module}\label{fig:hourglass}	
	\vspace{-0.5cm}
\end{figure}

\textbf{Density-aware propagation:}
{Since the context of neighboring people is important for crowd categorization (e.g., people waiting in line), we aim to encourage features to propagate mainly from person to person {\em along high-density regions.}}
Therefore, we propose a density-aware feature propagation method, 
where features propagating through low-density regions (background) decay faster than those in high-density regions (people).
 First, a dampening weight matrix $W_d\in \mathbb{R}^{h\times w}$ is defined based on the first-stage prediction of the density map $\hat{Y}' \in \mathbb{R}^{h\times w}$,
\begin{align}
W_d = \mathbbm{1}(\hat{Y}' \geq \epsilon)+\lambda \mathbbm{1}(\hat{Y}'<\epsilon),
\end{align}
where $\lambda \geq 0$ is the dampening factor and $\epsilon$ is a density threshold. Second, $W_d$ is used to decay features during propagation: 
 for CRF, the message is multiplied by  $W_d$ before being passed to other pixels; for GCN and Stacked Hourglass, the features are multiplied by $W_d$  before {each GCN layer or hourglass module}. 
{In this way, features are propagated along high-density regions, and decay exponentially when moving through low-density regions.}

\subsection{Complementary attention}
To exploit the complementary information between the two branches, we propose a complementary attention model. First, the first-stage density map and segmentation map are generated. Next, the first-stage segmentation features and the first-stage density map are concatenated, and then fed into the refinement framework. The first-stage density map serves as the attention for the segmentation branch,
which the model 
during feature propagation. Similarly, the refined segmentation map is concatenated with the first-stage counting features and serves as the attention for the counting refinement (see Fig.~\ref{fig:pipeline}).



\begin{figure}[t]
	\centering
	\footnotesize
	\begin{tabular}{@{}c@{}c@{}}
		(a) dampening factor  &  (b) the number of iterations \\
		\includegraphics[width=0.24\textwidth]{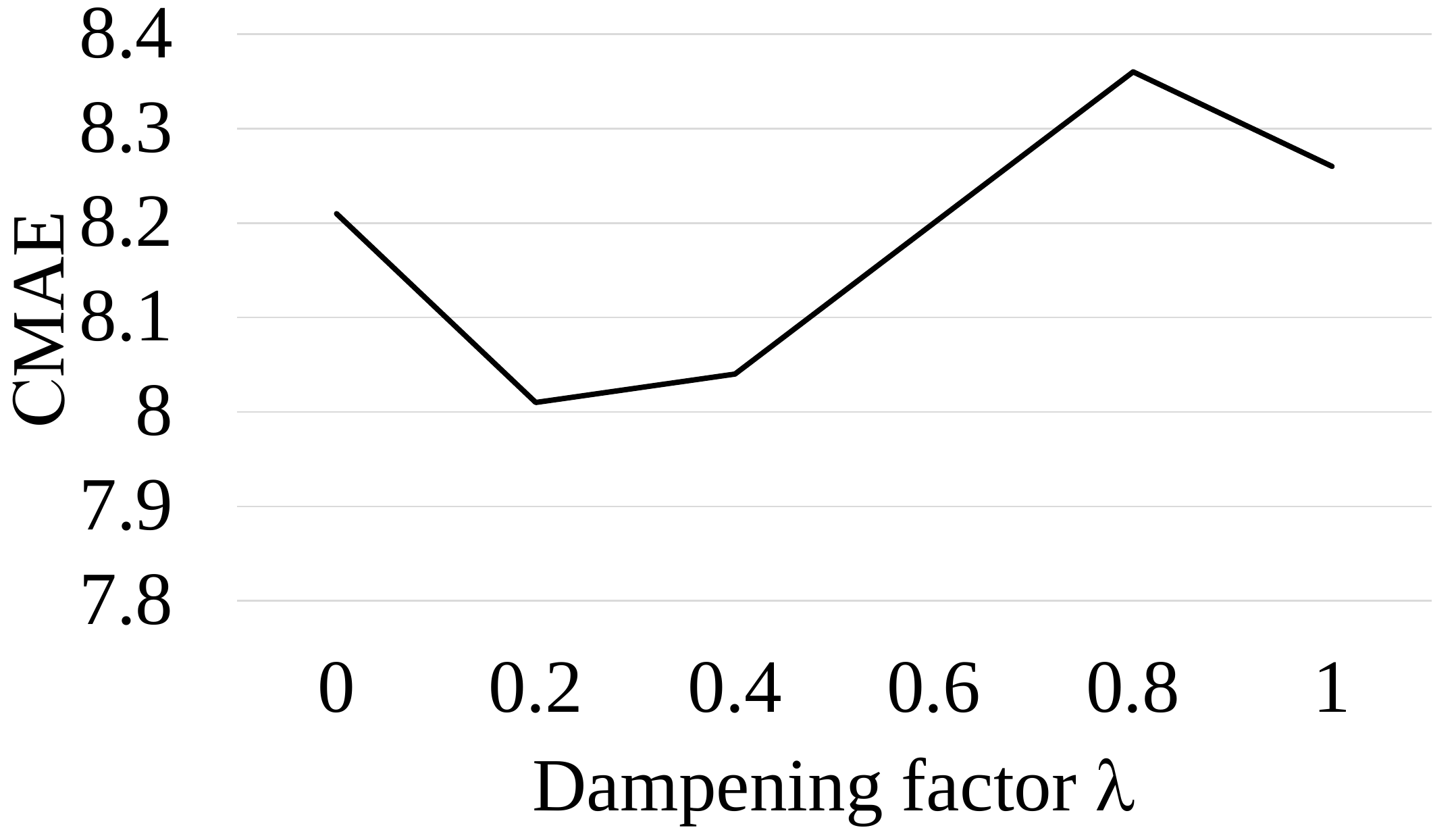} &
		\includegraphics[width=0.24\textwidth]{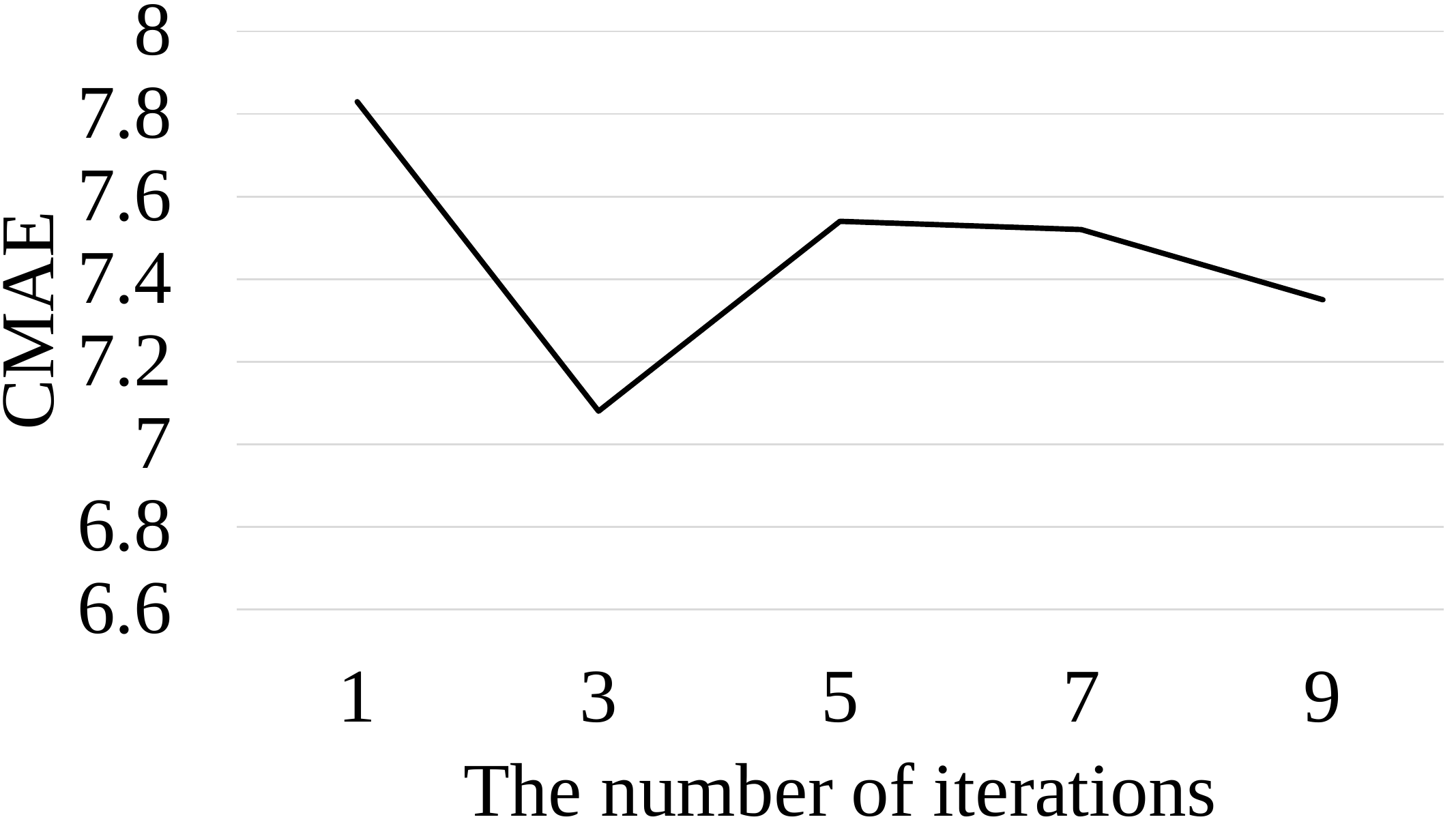}
	\end{tabular}
	\caption{\small CMAE versus (a) dampening factor $\lambda$, and (b) the number of iterations for density-aware feature propagation} 
	\label{fig:d_iter}
\end{figure}

\CUT{
\begin{table*}[t]
	\small
	\begin{center}
		\begin{tabular}{ l | c  c  c | c c c} \hline
			& Standing & Sitting   & CMAE & OMAE & Accuracy \% & Recall \% \\ \hline
			{No feature propagation}  &10.55&9.10&9.83&12.07&73.63&80.18, 50.02 \\
			GCN &10.35&8.42&9.38&12.31&73.62&80.73, 49.58 \\
			CRFs &10.02&\textbf{6.76}&8.39&10.62&78.76&77.37, 73.82 \\
			Hourglass (ours)  &\textbf{8.79}&7.23&\textbf{8.01}&9.28&76.48&78.23, 66.06 \\
			\hline 		
		\end{tabular}
	\end{center}
	\vspace{-0.2cm}
		\caption{Experiment results on the Standing/Sitting counting task using different feature propagation methods.
		The fine-grained counting performance is evaluated using Mean Absolute Error (MAE),
		and Category-averaged Mean Absolute Error (CMAE).
		The density map branch is evaluated using overall MAE (OMAE), and  the segmentation branch is evaluated using accuracy and recall.
}\label{tab:prop}
	\vspace{-0.5cm}
\end{table*}
}

\subsection{Loss functions}
The output of the density map estimation branch is the overall density map for all people, which is multiplied by the category segmentation maps  to obtain the fine-grained density maps. 
We use three loss functions during training: counting loss, segmentation loss and fine-grained loss.
Counting loss measures the mean square error (MSE) of the overall density map with the predicted density map,
	\begin{align}
	\ell_c= \|\hat{Y} - \sum_{j=1}^K Y_j\|^2+\|\hat{Y}'-\sum_{j=1}^k Y_j\|^2,
	\end{align}
where $Y_j$ is the GT density map of the $j$th category, and $\hat{Y}'$, $\hat{Y}$ are the first-stage (before the attention block) and final predicted density maps.
%
%
Segmentation loss measures the category segmentation error via soft cross entropy (SCE), 
\begin{align}
\ell_s= \sum_{j=1}^{K+1} -S_j \log \hat{S}'_{j}  - S_j \log \hat{S}_j,
\end{align}
where $S_j$ is the ground-truth segmentation map for the $j$th category, and $\hat{S}'_j$, $\hat{S}_j$ are the first-stage and final predicted segmentation maps.
%
Note that the segmentation maps are soft maps.
Fine-grained loss is a category-specific loss that measures the fine-grained counting error for each category,
\begin{align}
\ell_f= \sum_{j=1}^K \|\hat{Y}_j-Y_j\|^2_2.
\end{align}
where $\hat{Y}_j$ is the predicted density map for the $j$th category.
%
%
The final loss is computed by combining the three losses together,
\begin{equation}
l = l_c + \alpha l_s + \beta l_f,
\end{equation}
where $\alpha$ and $\beta$ are hyper-parameters.


\begin{table*}[t]
	\tiny
	\caption{ Ablation study on the Standing/Sitting task comparing: a) feature propagation methods; b) co-attention (CoAtt) and density-aware propagation (DAProp); c) combinations of loss functions.
		The fine-grained counting performance is evaluated using mean absolute error of each category (Standing/Sitting),
		and Category-averaged Mean Absolute Error (CMAE).
		The density map branch is evaluated using overall MAE (OMAE), and  the segmentation branch is evaluated using accuracy and recall.  The two recall values are for the two classes.
}\label{tab:ablationcombo}
	\begin{center}
	\resizebox{18cm}{!}{
		\begin{tabular}{ ll | c  c  c | c c c} \hline
			&& Standing & Sitting   & CMAE & OMAE & Accuracy \% & Recall \% \\ \hline
			\multirow{4}{*}{(a)}
			&{No feature propagation}  &10.55&9.10&9.83&12.07&73.63&80.18, 50.02 \\
			&GCN &10.35&8.42&9.38&12.31&73.62&80.73, 49.58 \\
			&CRF &10.02&\textbf{6.76}&8.39&10.62&78.76&77.37, 73.82 \\
			&Hourglass (ours)  &\textbf{8.79}&7.23&\textbf{8.01}&9.28&76.48&78.23, 66.06 \\
			\hline\hline
			\multirow{5}{*}{(b)}
			&Context  &10.20&\textbf{7.12}&8.66&9.73&77.84&77.84, 68.35 \\
			&Context + CoAtt  &9.48&7.04&8.26&9.18&78.70&81.33, 67.30 \\
			&Context + DAProp  &\textbf{8.39}&8.37&8.38&9.88&77.44&76.49, 73.24 \\
			&Context + NaiveAtt + DAProp &9.31&8.56&{8.94}&11.00&78.05&83.86, 58.82 \\
			&Context + CoAtt + DAProp (ours) &8.79&7.23&\textbf{8.01}&9.28&76.48&78.23, 66.06 \\
			\hline 		\hline
			\multirow{5}{*}{(c)}
			&FineGrained  &9.84&{7.55}&8.69&11.20&76.76&80.56, 61.44 \\
			&FineGrained + Seg  &9.80&7.62&8.71&10.24&77.36&77.94, 67.37 \\
			&FineGrained + Count  &9.99&8.24&9.11&9.72&74.45&73.70, 67.21 \\
			&Seg + Count  &11.84&8.82&10.33&9.41&78.85&79.48, 72.29 \\
			&FineGrained + Seg + Count (ours) &\textbf{8.79}&\textbf{7.23}&\textbf{8.01}&9.28&76.48&78.23, 66.06 \\
			\hline 		
		\end{tabular}
		}
	\end{center}

\end{table*}

\begin{figure*}[t]
	\begin{center}
		\includegraphics[width=\linewidth]{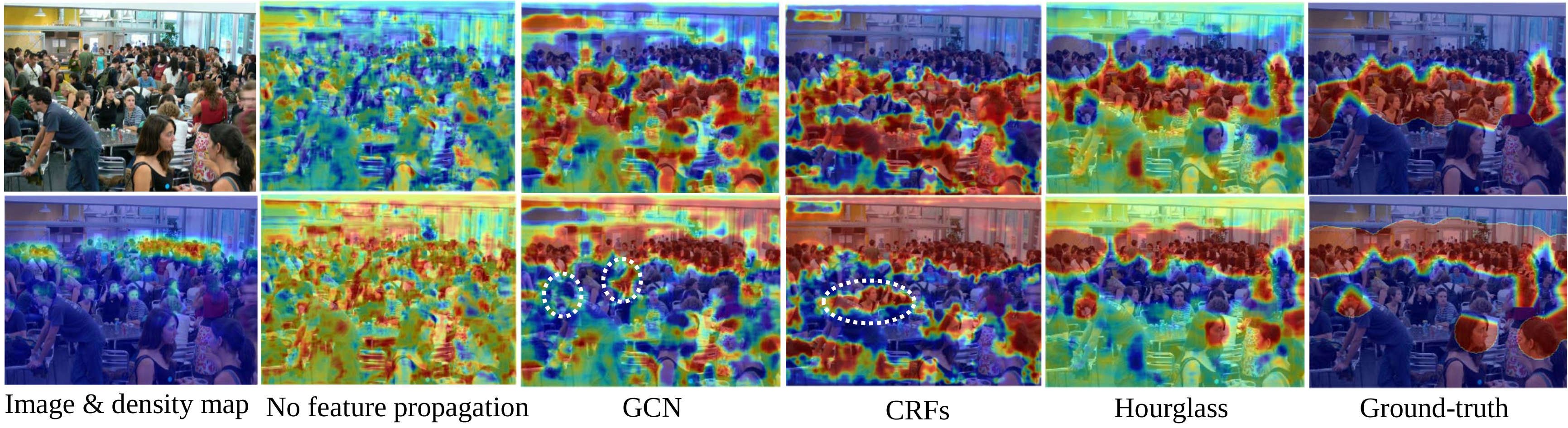}
	\end{center}
	\caption{\small The visualization of segmentation maps generated with and without feature propagation.
	{The first column shows the original image and first-stage predicted density map, while the last column shows the ground-truth segmentation for each category.}
		}\label{fig:seg_map}

\end{figure*}


\CUT{
\begin{table*}[t]	
	\begin{center}
	\small
		\begin{tabular}{ l | c  c  c |  c c c} \hline
			& Standing & Sitting   & CMAE & OMAE & Accuracy \% & Recall \% \\ \hline
			Context  &10.20&\textbf{7.12}&8.66&9.73&77.84&77.84, 68.35 \\
			Context + CoAtt  &9.48&7.04&8.26&9.18&78.70&81.33, 67.30 \\
			Context + DAProp  &\textbf{8.39}&8.37&8.38&9.88&77.44&76.49, 73.24 \\
			Context + NaiveAtt + DAProp &9.31&8.56&{8.94}&11.00&78.05&83.86, 58.82 \\
			Context + CoAtt + DAProp (Ours) &8.79&7.23&\textbf{8.01}&9.28&76.48&78.23, 66.06 \\
			\hline 		
		\end{tabular}
	\end{center}
		\caption{Experiment results on co-attention (CoAtt) and density-aware propagation (DAProp).}\label{tab:att_prop}
\end{table*}
}

\begin{figure}[t]
	\begin{center}
		\includegraphics[width=1.\linewidth]{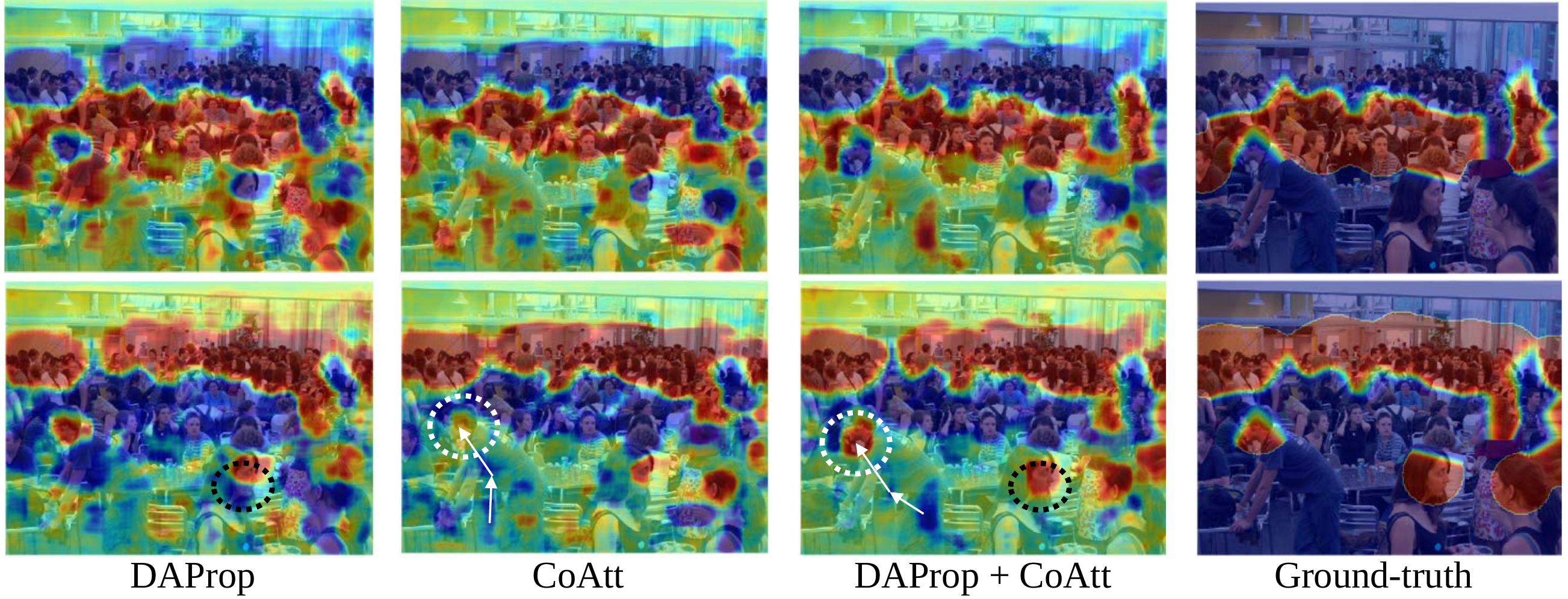}
	\end{center}
	\vspace{-0.5cm}
	\caption{The visualization of segmentation map generated with different components
	}\label{fig:att_prop}
	\vspace{-0.5cm}
\end{figure}

\section{Experiments}
\label{text:experiments}
We demonstrate the effectiveness of the proposed method through experiments on the new fine-grained counting dataset, including an ablation study on the model components as well as comparisons with baseline methods.

\subsection{Experiment Setup}
For the Towards/Away task, 
800 images are used for training and 1200 images for testing following the train/test split of the UCSD dataset \cite{4587569}. 
For the other 3 tasks, 100 images are used for testing and the remaining images used for training and validation. The learning rate is set to 0.0001 during training. For density-aware feature propagation, 
the dampening factor is $\lambda=0.2$, the density threshold is $\epsilon=0.001$,
and 3 iterations of propagation are used, which were determined by an initial experiment (see Fig.~\ref{fig:d_iter}).
The hyperparameters of the final loss are $\alpha=100$ and $\beta=10$.

The count predictions using the fine-grained density maps are evaluated using mean absolute error (MAE), for each category $j$,
	\begin{align}
	MAE_j = \frac{1}{n}\sum_{i=1}^n \left|\ssum(Y_j^i)-\ssum(\hat{Y}_j^i) \right|,
	\end{align}
where $Y_j^i$, $\hat{Y}_j^i$ are the ground-truth and predicted density maps for the $i$th test image and the $j$th category,  $n$ is the number of test images, and $\ssum$ is the spatial sum over the map.
The fine-grained counting performance is summarized with the category-averaged MAE (CMAE), 
    \begin{equation}
    CMAE = \frac{1}{K}\sum_{j=1}^K MAE_j,
    \end{equation}
We also evaluate the predictions from each branch of the architecture.
The count prediction of the overall density map is evaluated using overall MAE (OMAE),
	\begin{align}
	OMAE =  \frac{1}{n}\sum_{i=1}^n \left|\ssum(\sum\nolimits_{j=1}^K Y_j^i)-\ssum(\hat{Y}^i) \right|.
	\end{align}
The performance of the segmentation branch is evaluated using segmentation accuracy and recall {of the non-background pixels}.
\CUT{
{We ignore the background pixels when we calculate the accuracy:
	\begin{equation}
	\small
		Accuracy=\frac{1}{n}\sum_{i=1}^{n}\frac{\ssum(\mathbbm{1}(argmax (S^i)= argmax(\hat{S^j}) ))}{\ssum(\mathbbm{1}(\sum_{j=1}^{K}S^i_j))}
	\end{equation}
	where $S^i\in \mathbb{R}^{K\times h\times w}$, $\hat{S}^i\in \mathbb{R}^{K\times h\times w}$ are the ground-truth and predicted segmentation maps for $i$th image with background channel removed. And, the recall of category $j$ is given by:
	\begin{equation}\small
		recall_j = \frac{1}{n}\sum_{i=1}^{n}\frac{\ssum(\mathbbm{1}(argmax(S^i)=j \& argmax(\hat{S}^i)=j))}{\mathbbm{1}(argmax(S^i)=j)}
	\end{equation}
}
}
{Note that we do not evaluate segmentation precision since the false-negatives will be ignored due to the corresponding low values in the density map.} {Furthermore, the accuracy and recall are only used to roughly estimate the segmentation performance, which is less important compared to CMAE since the segmentation ground-truth is not accurate.}


\CUT{
\begin{table*}[t]	
	\begin{center}
	\small
		\begin{tabular}{ l | c  c  c |  c c c} \hline
			& Standing & Sitting   & CMAE & OMAE & Accuracy \%  & Recall \% \\ \hline
			FineGrained  &9.84&{7.55}&8.69&11.20&76.76&80.56, 61.44 \\
			FineGrained + Seg  &9.80&7.62&8.71&10.24&77.36&77.94, 67.37 \\
			FineGrained + Count  &9.99&8.24&9.11&9.72&74.45&73.70, 67.21 \\
			Seg + Count  &11.84&8.82&10.33&9.41&78.85&79.48, 72.29 \\
			FineGrained + Seg + Count &\textbf{8.79}&\textbf{7.23}&\textbf{8.01}&9.28&76.48&78.23, 66.06 \\
			\hline 		
		\end{tabular}
	\end{center}
		\caption{Experimental results of using different combinations of loss functions.}\label{tab:loss}
\end{table*}
}

\begin{table*}[t]	
	\begin{center}
	\caption{Experimental results comparing our method with crowd counting baselines on four fine-grained crowd counting tasks}\label{tab:baselines}
		\resizebox{18cm}{!}{
			\begin{tabular}{ l c | c  c c | c c c | c c c | ccc } \hline 
				& & \multicolumn{3}{c}{Standing/Sitting} & \multicolumn{3}{|c}{Waiting/Not waiting}& \multicolumn{3}{|c}{{Towards/Away}} & \multicolumn{3}{|c}{Violent/Non-violent} \\ 
				& & Standing & Sitting & CMAE & Waiting & Not waiting & CMAE &  Towards & Away & CMAE &  Violent & Non-violent & CMAE  \\ 
				\hline
				\multirow{3}{*}{OneNet}  
				& Fix4     & 13.89 & 10.00 & 11.95  & 5.34 & 3.72 & 4.53 & 2.38 & 3.76 & 3.07 &{4.33}&4.94&4.64  \\ 
				& Fix16    & 15.35 & 9.70 & 12.53  & 4.33 & 3.66 & 4.00 & 2.06 & 4.75 & 3.41 &4.97&4.88&4.93 \\ 
				& Adaptive & 12.14 & 9.14 & 10.64  & 4.64 & 3.93 & 4.28 & 1.89 & 4.28 & 3.09 &\textbf{4.26}&5.63&4.94 \\ \hline
				\multirow{3}{*}{TwoNets}  
				& Fix4     & 11.65 & 9.06 & 10.35 & 7.52 & 4.70 & 6.11 & 2.03 & 4.19 & 3.11 &4.99&5.21&5.10 \\ 
				& Fix16    & 12.62 & 9.34 & 10.98 & 4.97 & 4.00 & 4.49 & 2.11 & 5.13 & 3.62 &{4.34}&5.16&4.75 \\ 
				& Adaptive & 12.59 & 9.06 & 10.82 & 7.26 & 5.13 & 6.19 & 1.98 & 4.04 & {3.01} &4.50&4.91&4.70 \\ \hline
				\multirow{3}{*}{Segment}
				& Fix4  &  {11.09} & 9.15 & {10.12} & 3.78 & 3.78 & 3.78 & \underline{1.80} & {3.93} & \underline{2.86} &\underline{4.30}&\underline{4.54}&\underline{4.42}  \\  
				& Fix16 &  \underline{10.55} & {9.10} & \underline{9.83} & \underline{3.53} & \underline{3.54} & \underline{3.53}  & 3.06 & \textbf{3.16} & 3.11 &5.21&5.66&5.43 \\ 
				& Adaptive & 11.82 & \underline{9.04} & 10.43 & 3.93 & 3.90 & 3.91   & 2.08 & 4.04 & 3.06 &4.89&5.10&5.00  \\ \hline 
				Ours 
				&  & \textbf{8.79} & \textbf{7.23} & \textbf{8.01} & \textbf{2.88} & \textbf{3.10} & \textbf{2.99}  &\textbf{1.61}&\underline{3.38}&\textbf{2.49} &4.47&\textbf{4.23}&\textbf{4.35} \\  
				\hline
			\end{tabular}
		}
	\end{center}
\end{table*}

\subsection{Ablation study}

We first conduct an ablation study to show the efficacy of our model components.
{The ablation study uses the Standing/Sitting task of the dataset.}

{\bf Context:}
We first compare the three feature propagation methods with contextual information in Table \ref{tab:ablationcombo}(a).
 The counting performance is improved when using feature propagation, compared to when no feature propagation is used.
 This confirms that the context is important to distinguish between fine-grained crowd categories. 
 Comparing the three propagation methods, 
 GCN does not perform well
  because the construction of the graph using feature similarity is not effective.
  CRF performs better than GCN but less effective than the Stacked Hourglass, most likely because we use location and pixel color as hand-crafted features for message passing. However, these features may be too similar for people in different categories. Finally, stacked hourglass can effectively capture the global and local context by learning, and achieves the beset performance.

To better understand the effectiveness of contextual information, we visualize the segmentation maps generated by different approaches in Fig.~\ref{fig:seg_map}. Without contextual information (no feature propagation), the segmentation map is noisy as it is difficult to classify fine-grained categories by appearance features only. For GCN and CRF, the quality of the segmentation maps is improved. However, some areas are misclassified as circled in Fig.~\ref{fig:seg_map}. The segmentation maps generated by Stacked Hourglass are most similar to the ground-truth.

{\bf Density-aware propagation:}
We next evaluate the effectiveness of using the density map to guide feature propagation. From Table \ref{tab:ablationcombo}(b), our method using density-aware propagation (``Context+CoAtt+DAProp'') achieves better performance than ``Context+CoAtt'', and 
``Context+DAProp'' is better than ``Context'', which confirms 
density-aware feature propagation is effective. 

We also compare segmentation maps with or without DAProp in Fig.~\ref{fig:att_prop}. 
Without DAProp, the CoAtt model is less confident about the classification of the white circled area because it will be affected by background pixels during propagation (white arrows). However, with DAProp, the model is more confident about the classification result since the background features decay faster in low-density area (i.e. background non-crowd).

{\bf Complementary attention:}
Next we investigate the complementary attention module (see Table \ref{tab:ablationcombo}(b)).  Our method (``Context+CoAtt+DAProp'') outperforms ``Context+DAProp'',  and ``Context+CoAtt'' is better than ``Context'', which confirms the effectiveness of the complementary attention module. 
We also compare the proposed attention with naive attention (NaiveAtt), which directly multiplies the first-stage features with the attention map before refinement.
Replacing CoAtt 
 with NaiveAtt leads to worse OMAE performance, due to the loss of information after the NaiveAtt mechanism.
This confirms the counting branch also benefits from the context information from the segmentation branch.

In Fig.~\ref{fig:att_prop}, we compare the segmentation maps between methods with and without CoAtt. As circled in black, the human face is incorrectly segmented without CoAtt. In contrast, the same face is segmented more accurately when using CoAtt 
since the model can focus on high-density areas.

{\bf Loss functions:}
Finally, we evaluate different combinations of the loss functions: fine-grained loss (FineGrained), segmentation  (Seg), and counting loss (Count). 
The results of various combinations are presented in Table \ref{tab:ablationcombo}(c). Fine-grained loss is the most important loss function since the performance significantly decreases without it. The best performance is achieved by combining these three loss functions together. 
{Finally, both Accuracy and Recall of the proposed method (``FineGrained+Seg+Count'') outperform ``FineGrained+Count'', which shows that the segmentation loss is effective to improve the segmentation quality.}


\subsection{Comparison with counting baselines}
To confirm the effectiveness of our proposed method, we compare with four baseline methods:

\emph{OneNet}: directly predict all the density maps for the categories with one network.  {Each category corresponds to one output channel of the network, and the feature maps are shared among the categories.} 

\emph{TwoNets}: use a separate network for each category, {i.e., no feature sharing between categories.}

\emph{Segment}: 
		{Use a shared feature extractor and predict an overall density map and a segmentation map (without using contextual information), which is equivalent to the first-stage prediction of our method.}
		{The category density maps are obtained by element-wise multiplying the  density map with the segment map.}

\CUT{
\emph{Detection}: use an object detector to detect people, where each crowd category is a separate object class.
	{Since bounding boxes are not labeled in the  dataset, we first generate a ground-truth bounding box on each dot annotation 
 using a tiny face detector \cite{yoo2019extd} with low threshold 0.2.  The class label is determined from the category density in the bounding box. Then, a detector  \cite{yoo2019extd}  for the crowd categories is fine-tuned on the training set with the generated ground-truth bounding boxes. The category crowd count is obtained by counting the number of detections for each category.}
}

The  results are presented in Table \ref{tab:baselines}. 
{We test different ground-truth density maps, generated via fixed bandwidth (4 and 16, denoted as Fix4 and Fix16), and adaptive bandwidth (Adaptive). Due to differences in image resolution between the tasks, Segment and Ours perform better using Fix16 on tasks with high resolution images (Standing/Sitting, Waiting/Not waiting), and  Fix4 on tasks with low resolution images (Towards/Away, Violent/Non-violent).  We present results using these settings for Ours.}

The segmentation-based approaches (Ours, Segment) outperform direct prediction with one or two networks (OneNet, TwoNets), since the segmentation branch 
{provides more meaningful supervision of the different categories, compared to a network learning a mix of density and category at the same time.}
%
Furthermore, our proposed method is consistently better than Segment (the first-stage prediction), which validates the usefulness of context information and information sharing between branches for solving the fine-grained counting task. {Examples of fine-grained crowd counting results for different methods are shown in Fig. \ref{fig:examples}.}

\begin{figure*}
	\begin{center}
		\includegraphics[width=\linewidth]{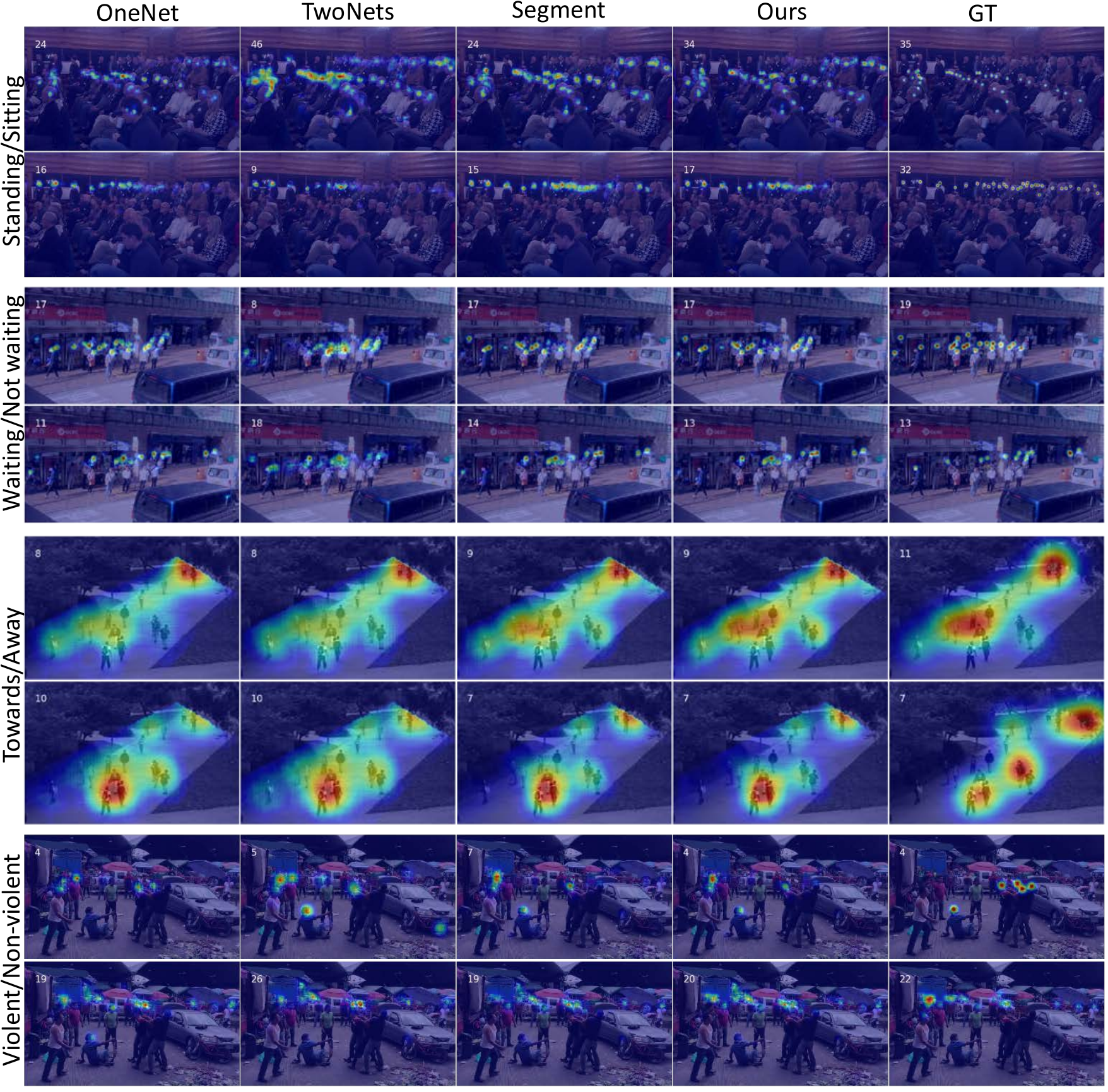}
	\end{center}
	\caption{Examples of fine-grained crowd counting results for different approaches on 4 tasks. The number in the top-left of the image is the predicted or ground-truth count.}\label{fig:examples}
\end{figure*}

 


\subsection{Comparison of counting backbones}


We also run an experiment with a stronger backbone -- CSRNet which is pre-trained ImageNet \cite{deng2009imagenet}. Table \ref{tab:csrnet} shows the results using CSRNet for fine-grained counting (either one network or two networks), and our method using CSRNet as the backbone: {1) with joint-training of the density map and segmentation networks (Ours); 2) only training the segmentation network (w/o joint train); 3) fine-tuning the CSRNet backbone and training the segmentation network separately (finetune w/o joint train).}
Comparing backbones for our method, the CMAE values are generally lower using the deeper pre-trained CSRNet than with the shallower FCN-7. Comparing methodologies, the results using the CSRNet backbone are consistent with those using the FCN-7 backbone -- our method using context outperforms the simple one-net and two-net versions. {Also, our method with joint training achieves better performance than methods without joint training, which confirms the effectiveness of training the segmentation and counting branches together.}

\begin{table*}[]	
	\begin{center}
		\caption{Experimental results comparing our method with CSRNet}\label{tab:csrnet}
		\resizebox{18cm}{!}{
			\begin{tabular}{ l  | c  c c | c c c | c c c | ccc } \hline 
				& \multicolumn{3}{c}{Standing/Sitting} & \multicolumn{3}{|c}{Waiting/Not waiting}& \multicolumn{3}{|c}{{Towards/Away}} & \multicolumn{3}{|c}{Violent/Non-violent} \\ 
				& Standing & Sitting & CMAE & Waiting & Not waiting & CMAE &  Towards & Away & CMAE &  Violent & Non-violent & CMAE  \\ \hline
				{CSRNet}  
				& 8.42 & 7.11 & 7.77  & 2.84 & 2.89 & 2.87 & 2.55 & 2.83 & 2.69 & 4.21 & 3.52 & 3.87 \\ 
				{TwoCSRNets}  
				& 8.68 & 8.07 & 8.38 & 3.19 & 2.84 & 3.02 & 2.91 & 3.05 & 2.98 & \textbf{3.82} & 4.19 & 4.01 \\ 
				
				CSRNet + seg (w/o joint train) &42.23&12.61&27.42&12.50&8.74&10.62&8.78&12.47&10.61&5.60&7.34&6.49\\ 
				CSRNet + seg + finetune (w/o joint train) &9.09&6.06&7.58&2.68&2.72&2.70&1.41&\textbf{1.84}&1.63&4.37&3.82&4.10\\ \hline
				Ours (CSRNet, w/ joint train)
				& \textbf{8.36} & \textbf{5.56} & \textbf{6.96} & \textbf{2.61} & \textbf{2.67} & \textbf{2.64} & \textbf{1.30} & {1.89} & \textbf{1.60} & 4.34 & \textbf{3.32} & \textbf{3.83} \\    
				\hline
				
			\end{tabular}
		}
	\end{center}
\end{table*}

\subsection{Comparison with detection}
We test two detection algorithms for fine-grained crowd counting: TinyFace \cite{yoo2019extd} which detects human face and YOLOv3 \cite{redmon2018yolov3} which detects human body. We fine-tune the pre-trained detectors to detect people from different categories.
TinyFace is pre-trained on the WIDER Face dataset \cite{yang2016wider} and YOLOv3 is pre-trained on COCO \cite{lin2014microsoft}. Since bounding boxes are not labeled in our proposed dataset, we first generate ground-truth bounding boxes on dot annotations using the pre-trained detectors (TinyFace or YOLOv3) with a low threshold of 0.2. The class label is determined from the category density in the bounding box. Then, a detector for the crowd categories is fine-tuned on the fine-grained counting training set with the generated ground-truth bounding boxes. The category count is obtained by counting the number of detections for each category.

The experimental results using detector-based counting are shown in Table \ref{tab:det}. TinyFace outperforms YOLOv3 on Standing/Sitting task because it is more reliable to detect faces than the whole human body in crowded scenes. For the other three tasks, which are less crowded than Standing/Sitting, the human body detector YOLOv3 is better than TinyFace. However, both detection methods are worse than our proposed approach for fine-grained counting.
{The detection approaches do not work well, compared to the density map approaches,  because of the low appearance variations between crowd categories and need for long-range context.}

In Table \ref{tab:omae}, we also compare the overall counting error (OMAE) of detectors and our approach. TinyFace achieves better overall counting performance than YOLOv3 on high density scenes, such as the Standing/Sitting and Waiting/Not-waiting tasks, because face detector works better for occluded scenes. For low density images from tasks like Violent/Non-violent, the YOLOv3 human body detector is better. The overall counting performance of our proposed method is significantly better than detection based approaches for all tasks.

We also visualize detection results of two detectors on the four fine-grained counting tasks in Figure \ref{fig:det}. For crowded scenes like Standing/Sitting and Waiting/Not-waiting, TinyFace can detect most of the people in images, while the YOLOv3 body detector misses many people far away from camera or occluded by other people. For applications with less people, the YOLOv3 body detector is better than a face detector because more contextual information can be exploited.
Both detectors achieve bad performance for fine-grained crowd counting because it is difficult to distinguish fine-grained categories only by appearance features. As shown in Figure \ref{fig:det}, some of the detections are classified as both categories. This confirms that it is difficult to solve fine-grained crowd counting by detection based approaches.

\begin{table*}
	\begin{center}
		\caption{Experimental results comparing our method with detection baselines on four fine-grained crowd counting tasks 
		}\label{tab:det}
		\resizebox{18cm}{!}{
			\begin{tabular}{ l | c  c c | c c c | c c c | ccc } \hline 
				& \multicolumn{3}{c}{Standing/Sitting} & \multicolumn{3}{|c}{Waiting/Not waiting}& \multicolumn{3}{|c}{{Towards/Away}} & \multicolumn{3}{|c}{Violent/Non-violent} \\ 
				& Standing & Sitting & CMAE & Waiting & Not waiting & CMAE &  Towards & Away & CMAE &  Violent & Non-violent & CMAE  \\ \hline
				TinyFace \cite{yoo2019extd} &12.17&11.25&11.71&3.60&7.56&5.58&9.33&16.42&12.88&7.32&7.88&7.60 \\ 
				YOLOv3 \cite{redmon2018yolov3} &37.29&11.98&24.64&4.44&4.95&4.70&5.67&5.49&5.58&4.74&6.53&5.64 \\ \hline
				Ours 
				&  \textbf{8.79} & \textbf{7.23} & \textbf{8.01} & \textbf{2.88} & \textbf{3.10} & \textbf{2.99}  &\textbf{1.61}&\textbf{3.38}&\textbf{2.49} &\textbf{4.47}&\textbf{4.23}&\textbf{4.35} \\  \hline
			\end{tabular}
		}
	\end{center}
\end{table*}

\begin{table*}	
	\begin{center}
		\small
		\caption{Comparison of overall counting performance (OMAE) between detection approaches and our proposed method
		}\label{tab:omae}
		
			\begin{tabular}{ l | c  c c c } \hline 
				& {Standing/Sitting} & {Waiting/Not waiting}& {{Towards/Away}} & {Violent/Non-violent} \\  \hline
				TinyFace \cite{yoo2019extd} &13.64&6.58&25.75&15.20 \\ 
				YOLOv3 \cite{redmon2018yolov3} &48.87&8.41&11.01&10.52 \\ \hline
				Ours 
				&  \textbf{9.66} & \textbf{3.29} & \textbf{3.64} & \textbf{6.34}  \\  \hline
			\end{tabular}
	\end{center}
\end{table*}

\begin{figure*}
	\begin{center}
		\includegraphics[width=\linewidth]{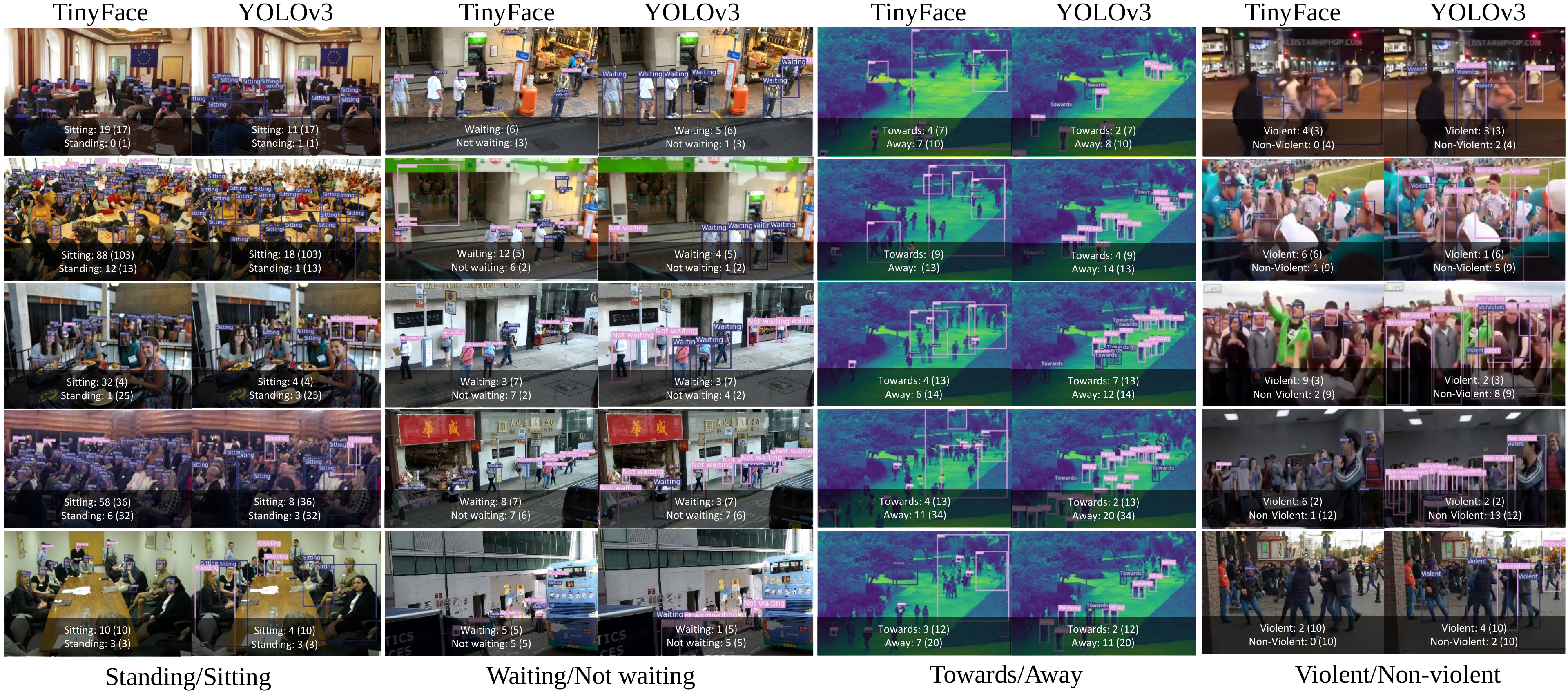}
	\end{center}
	\caption{Examples of detection results on fine-grained crowd counting. {The numbers in parentheses are the ground-truths counts.}
	}\label{fig:det}
\end{figure*}

\section{Conclusion}
\label{text:conc}
In this paper, we propose the problem of fine-grained crowd counting which divides people in a crowd into categories based on low-level behaviors (e.g., facing direction, pose, violent actions, and waiting in line) and counts the number of people in each category.
{The key point of fine-grained crowd counting is how to utilize context to distinguish different behaviours, which is scientifically different from crowd counting.}
 To promote research on fine-grained crowd counting, we construct a new dataset that contains images representing four fine-grained counting tasks. Finally, a two branch architecture consisting of density map and segmentation branches, and utilizing density-aware feature propagation and complementary attention to refine predictions, is proposed to solve fine-grained counting.
 Extensive experiments on the four applications confirm the effectiveness of the proposed method.

\section*{Acknowledgements}
This work was supported by grants from the Research Grants Council of the Hong Kong Special Administrative Region, China (Project No. [T32-101/15-R] and CityU 11212518).

\bibliographystyle{IEEEtran}
\bibliography{egbib}

\end{document}